# Distributed Tree Kernels

**Fabio Massimo Zanzotto**  FABIO.MASSIMO.ZANZOTTO@UNIROMA2.IT

University of Rome Tor Vergata, Viale del Politecnico, 1, 00133 Rome, Italy

**Lorenzo Dell'Arciprete**  LORENZO.DELLARCIPRETE@GMAIL.COM

University of Rome Tor Vergata, Viale del Politecnico, 1, 00133 Rome, Italy

## Abstract

In this paper, we propose the *distributed tree kernels* (DTK) as a novel method to reduce time and space complexity of tree kernels. Using a linear complexity algorithm to compute vectors for trees, we embed feature spaces of tree fragments in low-dimensional spaces where the kernel computation is directly done with dot product. We show that DTKs are faster, correlate with tree kernels, and obtain a statistically similar performance in two natural language processing tasks.

## 1. Introduction

Trees are fundamental data structures used to represent very different objects such as proteins, HTML documents, or interpretations of natural language utterances. Thus, many areas – for example, biology (Vert, 2002; Hashimoto et al., 2008), computer security (Düssel et al., 2008), and natural language processing (Collins & Duffy, 2001; Gildea & Jurafsky, 2002; Pradhan et al., 2005; MacCartney et al., 2006) – fostered extensive research in methods for learning classifiers that leverage on these data structures.

Tree kernels (TK), firstly introduced in (Collins & Duffy, 2001) as specific convolution kernels (Haussler, 1999), are widely used to fully exploit tree structured data when learning classifiers. Different tree kernels modeling different feature spaces have been proposed (see (Shin et al., 2011) for a survey), but a primary research focus is the reduction of their execution time. Kernel machines compute many times TK functions during learning and classification. The original tree kernel algorithm (Collins & Duffy, 2001), that relies on dynamic programming techniques, has a quadratic time and space complexity with respect to the size of input trees. Execution time and space occupation are still affordable for parse trees of natural language sentences that hardly go beyond the hundreds of nodes (Rieck et al., 2010). But these tree kernels hardly scale to large training and application sets.

As worst-case complexity of TKs is hard to improve, the biggest effort has been devoted in controlling the average execution time of TK algorithms. Three directions have been mainly explored. The first direction is the exploitation of some specific characteristics of trees. For example, it is possible to demonstrate that the execution time of the original algorithm becomes linear in average for parse trees of natural language sentences (Moschitti, 2006). Yet, the tree kernel has still to be computed over the full underlying feature space and the space occupation is still quadratic. The second explored direction is the reduction of the underlying feature space of tree fragments to control the execution time by approximating the kernel function. The feature selection is done in the learning phase. Then, for the classification, either the selection is directly encoded in the kernel computation by selecting subtrees headed by specific node labels (Rieck et al., 2010) or the smaller selected space is made explicit (Pighin & Moschitti, 2010). In these cases, the beneficial effect is only during the classification and learning is overloaded with feature selection. The third direction exploits dynamic programming on the whole training and application sets of instances (Shin et al., 2011). Kernel functions are reformulated to be computed using partial kernel computations done for other pairs of trees. As any dynamic programming technique, this approach is transferring time complexity in space complexity.

In this paper, we propose the *distributed tree kernels* (DTK) (introduced in (Zanzotto & Dell'Arciprete, 2011)) as a novel method to reduce time and space complexity of tree kernels. The idea is to embed feature spaces of tree fragments in low-dimensional spaces, where the computation is approximated but its worst-case complexity is linear with respect to the dimension of the space. As a direct embedding is impractical, we propose a recursive algorithm with linear complexity to compute reduced vectors for trees in the low-dimensional space. We formally show that the dot product among reduced vectors approximates the original tree kernel when a vector composition function with specific ideal properties is used. We then propose two approximations of the ideal vector composition function and we study their properties. Finally, we empirically investigate the execution time of DTKs and how well these new kernels approximate original tree kernels. We show that DTKs are faster, correlate with tree kernels, and obtain a statistically similar performance in two natural language processing tasks.

The rest of the paper is organized as follows. Section 2 introduces the notation, the basic idea, and the expected properties for DTKs. Section 3 introduces the DTKs and proves their properties. Section 4 compares the complexity of DTKs with other tree kernels. Section 5 empirically investigates these new kernel algorithms. Finally, section 6 draws some conclusions.

## 2. Challenges for Distributed Tree Kernels

### 2.1. Notation and Basic Idea

Tree kernels (TK) (Collins & Duffy, 2001) have been proposed as efficient methods to implicitly compute dot products in feature spaces $\mathbb{R}^m$ of tree fragments. A direct computation in these high-dimensional spaces is impractical. Given two trees, $T_1$ and $T_2$ in $\mathbb{T}$, tree kernels $TK(T_1, T_2)$ perform weighted counts of the common subtrees $\tau$. By construction, these counts are the dot products of the vectors representing the trees, $\vec{T}_1$ and $\vec{T}_2$ in $\mathbb{R}^m$, i.e.:

$$TK(T_1, T_2) = \vec{T}_1 \cdot \vec{T}_2 \qquad (1)$$

Vectors $\vec{T}$ encode trees $T$ as *forests* of active tree fragments $\mathcal{F}(T)$. Each *dimension* $\vec{\tau}_i$ of $\mathbb{R}^m$ corresponds to a tree fragment $\tau_i$. The trivial weighting scheme assigns $\omega_i = 1$ to dimension $\vec{\tau}_i$ if tree fragment $\tau_i$ is a subtree of the original tree $T$ and $\omega_i = 0$ otherwise. Different weighting schemes are possible and used. Function $\mathcal{I}$, that maps trees in $\mathbb{T}$ to vectors in $\mathbb{R}^m$, is:

$$\vec{T} = \mathcal{I}(T) = \sum_{i=1}^{m} \omega_i I(\tau_i) = \sum_{i=1}^{m} \omega_i \vec{\tau}_i \qquad (2)$$

where $I$ maps tree fragments into related vectors of the standard orthogonal basis of $\mathbb{R}^m$, i.e., $\vec{\tau}_i = I(\tau_i)$.

To reduce computational complexity of tree kernels, we want to explore the possibility of embedding vectors $\vec{T} \in \mathbb{R}^m$ into smaller vectors $\widetilde{\vec{T}} \in \mathbb{R}^d$, with $d \ll m$, to allow for an approximated but faster and explicit computation of these kernel functions. The direct embedding $f : \mathbb{R}^m \to \mathbb{R}^d$ is, in principle, possible with techniques like singular value decomposition or random indexing (Sahlgren, 2005), but it is again impractical due to the huge dimension of $\mathbb{R}^m$.

Then, our basic idea is to look for a function $\widehat{F} : \mathbb{T} \to \mathbb{R}^d$ that directly maps trees $T$ into small vectors $\widetilde{\vec{T}}$. We call these latter *distributed trees* (DT) in line with Distributed Representations (Hinton et al., 1986). The computation of similarity over *distributed trees* is the *distributed tree kernel* (DTK):

$$DTK(T_1, T_2) \triangleq \widetilde{\vec{T}}_1 \cdot \widetilde{\vec{T}}_2 = \widehat{F}(T_1) \cdot \widehat{F}(T_2) \qquad (3)$$

As the two distributed trees are in the low dimensional space $\mathbb{R}^d$, the dot product computation, having constant complexity, is extremely efficient. Computation of function $\widehat{F}$ is more expensive than the actual DTK, but it is done once for each tree and outside of the learning algorithms. We also propose a recursive algorithm with linear complexity to perform this computation.

### 2.2. Distributed Trees, Distributed Tree Fragments, and Expected Properties

Distributed tree kernels are faster than tree kernels. We here examine the properties required of $\widehat{F}$ so that DTKs are also approximated computations of TKs, i.e.:

$$DTK(T_1, T_2) \approx TK(T_1, T_2) \qquad (4)$$

To derive these properties and describe function $\widehat{F}$, we show the relations between the traditional function $\mathcal{I} : \mathbb{T} \to \mathbb{R}^m$ that maps trees into forests of tree fragments, in the tree fragments feature space, $I : \mathbb{T} \to \mathbb{R}^m$ that maps tree fragments into the standard orthogonal basis of $\mathbb{R}^m$, the linear embedding function $f : \mathbb{R}^m \to \mathbb{R}^d$ that maps $\vec{T}$ into a smaller vector $\widetilde{\vec{T}} = f(\vec{T})$, and our newly defined function $\widehat{F}$.

Equation 2 presents vectors $\vec{T}$ with respect to the *standard orthonormal basis* $E = \{\vec{e}_1 \ldots \vec{e}_m\} = \{\vec{\tau}_1 \ldots \vec{\tau}_m\}$ of $\mathbb{R}^m$. Then, according to this reading, we can rewrite the distributed tree $\widetilde{\vec{T}} \in \mathbb{R}^d$ as:

$$\widetilde{\vec{T}} = f(\vec{T}) = f(\sum_i \omega_i \vec{\tau}_i) = \sum_i \omega_i f(\vec{\tau}_i) = \sum_i \omega_i \widetilde{\vec{\tau}}_i$$

where each $\widetilde{\vec{\tau}}_i$ represents tree fragment $\tau_i$ in the new space. The linear function $f$ works as a sort of *approximated basis transformation*, mapping vectors $\vec{\tau}$ of the standard basis $E$ into approximated vectors $\widetilde{\vec{\tau}}$ that should represent them. As $\widetilde{\vec{\tau}}_i$ represents a single tree fragment $\tau_i$, we call it a *distributed tree fragment* (DTF). The set of vectors $\widetilde{E} = \{\widetilde{\vec{\tau}}_1 \ldots \widetilde{\vec{\tau}}_m\}$ should be the *approximated orthonormal basis* of $\mathbb{R}^m$ embedded in $\mathbb{R}^d$. Then, these two properties should hold:

**Property 1** (Nearly Unit Vectors) *A distributed tree fragment $\widetilde{\vec{\tau}}$ representing a tree fragment $\tau$ is a nearly unit vector:* $1 - \epsilon < ||\widetilde{\vec{\tau}}|| < 1 + \epsilon$

**Property 2** (Nearly Orthogonal Vectors) *Given two different tree fragments $\tau_1$ and $\tau_2$, their distributed vectors are nearly orthogonal: if $\tau_1 \neq \tau_2$, then $|\vec{\widetilde{\tau}}_1 \cdot \vec{\widetilde{\tau}}_2| < \epsilon$*

As vectors $\vec{\widetilde{\tau}} \in \widetilde{E}$ represent the basic tree fragments $\tau$, the idea is that $\vec{\widetilde{\tau}}$ can be obtained directly from tree fragment $\tau$ by means of a function $\widehat{f}(\tau) = f(I(\tau))$ that composes $f$ and $I$. Using this function to obtain distributed tree fragments $\vec{\widetilde{\tau}}$, distributed trees $\vec{\widetilde{T}}$ can be obtained as follows:

$$\vec{\widetilde{T}} = \widehat{F}(T) = \sum_{\tau_i \in \mathcal{F}(T)} \omega_i \widehat{f}(\tau_i) \qquad (5)$$

This latter equation is presented with respect to the active tree fragments forest $\mathcal{F}(T)$ of $T$, neglecting vectors where $\omega_i = 0$. It is easy to show that, if properties 1 and 2 hold for function $\widehat{f}$, distributed tree kernels approximate tree kernels (see Equation 4).

## 3. Computing Distributed Tree Fragments and Distributed Trees

Johnson-Lindenstrauss Lemma (JLL) (Johnson & Lindenstrauss, 1984) guarantees that the embedding function $f : \mathbb{R}^m \to \mathbb{R}^d$ exists. It also points out the relation between the desired approximation $\epsilon$ of Property 2 (Nearly Orthogonal Vectors) and the required dimension $d$ of the target space, for a certain value of dimension $m$. This relation affects how well DTKs approximate TKs (Equation 4).

Knowing that $f$ exists, we are presented with the following issues:

- building a function $\widehat{f}$ that directly computes the distributed tree fragment $\vec{\widetilde{\tau}}_i$ from tree fragment $\tau_i$ (Sec. 3.2);
- showing that distributed trees $\vec{\widetilde{T}} = \widehat{F}(T)$ can be computed efficiently (Sec. 3.1).

Once the above issues are solved, we need to empirically show that Equation (4) is satisfied and that computing DTKs is more efficient than computing TKs. These latter points are discussed in the experimental section.

### 3.1. Computing Distributed Tree Fragments from Trees

This section introduces function $\widehat{f}$ for distributed tree fragments and shows that, using an ideal vector composition function $\diamond$, the proposed function $\widehat{f}(\tau_i)$ satisfies properties 1 and 2.

3.1.1. REPRESENTING TREES AS VECTORS

The basic blocks needed to represent trees are their nodes. We then start from a set $\mathcal{N} \subset \mathbb{R}^d$ of nearly orthonormal vectors representing nodes. Each node $n$ is mapped to a vector $\vec{\widetilde{n}} \in \mathcal{N}$. To ensure that these basic vectors are statistically nearly orthonormal, their elements $(\widetilde{n})_i$ are randomly drawn from a normal distribution $N(0, 1)$ and they are normalized so that $||\vec{\widetilde{n}}|| = 1$ (cf. the demonstration of Johnson-Lindenstrauss Lemma in (Dasgupta & Gupta, 1999)). Actual node vectors depend on the node labels, so that $\vec{\widetilde{n}}_1 = \vec{\widetilde{n}}_2$ if $\mathcal{L}(n_1) = \mathcal{L}(n_2)$, where $\mathcal{L}(\cdot)$ is the node label.

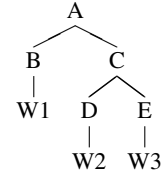

*Figure 1.* A sample tree

Tree structure can be univocally represented in a 'flat' format using a parenthetical notation. For example, the tree in Fig. 1 is represented by the sequence (A (B W1)(C (D W2)(E W3))). This notation corresponds to a depth-first visit of the tree, augmented with parentheses so that the tree structure is determined as well.

Replacing the nodes with their corresponding vectors and introducing a vector composition function $\diamond : \mathbb{R}^d \times \mathbb{R}^d \to \mathbb{R}^d$, the above formulation can be seen as a mathematical expression that defines a representative vector for a whole tree. The example tree would then be represented by vector $\vec{\widetilde{\tau}} = (\vec{\widetilde{A}} \diamond (\vec{\widetilde{B}} \diamond \vec{\widetilde{W1}}) \diamond (\vec{\widetilde{C}} \diamond (\vec{\widetilde{D}} \diamond \vec{\widetilde{W2}}) \diamond (\vec{\widetilde{E}} \diamond \vec{\widetilde{W3}})))$.

Then, we formally define function $\widehat{f}(\tau)$, as follows:

**Definition 1** *Let $\tau$ be a tree and $\mathcal{N}$ the set of nearly orthogonal vectors for node labels. We recursively define $\widehat{f}(\tau)$ as:*

- $\widehat{f}(n) = \vec{\widetilde{n}}$ *if $n$ is a terminal node, where $\vec{\widetilde{n}} \in \mathcal{N}$*

- $\widehat{f}(\tau) = (\vec{\widetilde{n}} \diamond \widehat{f}(\tau_{c_1} \ldots \tau_{c_k}))$ *if $n$ is the root of $\tau$ and $\tau_{c_i}$ are its children subtrees*

- $\widehat{f}(\tau_1 \ldots \tau_k) = (\widehat{f}(\tau_1) \diamond \widehat{f}(\tau_2 \ldots \tau_k))$ *if $\tau_1 \ldots \tau_k$ is a sequence of trees*

3.1.2. THE IDEAL VECTOR COMPOSITION FUNCTION

We here introduce the ideal properties of the vector composition function $\diamond$, such that function $\widehat{f}(\tau_i)$ has the two desired properties.

The definition of the ideal composition function follows:

**Definition 2** *The ideal composition function is $\diamond : \mathbb{R}^d \times \mathbb{R}^d \to \mathbb{R}^d$ such that, given $\vec{\widetilde{a}}, \vec{\widetilde{b}}, \vec{\widetilde{c}}, \vec{\widetilde{d}} \in \mathcal{N}$, a scalar $s$* *and a vector $\widetilde{\vec{t}}$ obtained composing an arbitrary number of vectors in $\mathcal{N}$ by applying $\diamond$, the following properties hold:*

*2.1 Non-commutativity with a very high degree $k$[1]*

*2.2 Non-associativity: $\widetilde{\vec{a}} \diamond (\widetilde{\vec{b}} \diamond \widetilde{\vec{c}}) \neq (\widetilde{\vec{a}} \diamond \widetilde{\vec{b}}) \diamond \widetilde{\vec{c}}$*

*2.3 Bilinearity:*

$$I) \ (\widetilde{\vec{a}} + \widetilde{\vec{b}}) \diamond \widetilde{\vec{c}} = \widetilde{\vec{a}} \diamond \widetilde{\vec{c}} + \widetilde{\vec{b}} \diamond \widetilde{\vec{c}}$$
$$II) \ \widetilde{\vec{c}} \diamond (\widetilde{\vec{a}} + \widetilde{\vec{b}}) = \widetilde{\vec{c}} \diamond \widetilde{\vec{a}} + \widetilde{\vec{c}} \diamond \widetilde{\vec{b}}$$
$$III) \ (s\widetilde{\vec{a}}) \diamond \widetilde{\vec{b}} = \widetilde{\vec{a}} \diamond (s\widetilde{\vec{b}}) = s(\widetilde{\vec{a}} \diamond \widetilde{\vec{b}})$$

*Approximation Properties*

*2.4 $||\widetilde{\vec{a}} \diamond \widetilde{\vec{b}}|| = ||\widetilde{\vec{a}}|| \cdot ||\widetilde{\vec{b}}||$*

*2.5 $|\widetilde{\vec{a}} \cdot \widetilde{\vec{t}}| < \epsilon$ if $\widetilde{\vec{t}} \neq \widetilde{\vec{a}}$*

*2.6 $|\widetilde{\vec{a}} \diamond \widetilde{\vec{b}} \cdot \widetilde{\vec{c}} \diamond \widetilde{\vec{d}}| < \epsilon$ if $|\widetilde{\vec{a}} \cdot \widetilde{\vec{c}}| < \epsilon$ or $|\widetilde{\vec{b}} \cdot \widetilde{\vec{d}}| < \epsilon$*

The ideal function $\diamond$ cannot exist. Property 2.5 can be only statistically valid and never formally as it opens to an infinite set of nearly orthogonal vectors. But, this function can be approximated (see Sec. 5.1).

### 3.1.3. PROPERTIES OF DISTRIBUTED TREE FRAGMENTS

Having defined the ideal basic composition function $\diamond$, we can now focus on the two properties needed to have DTFs as a nearly orthonormal basis of $\mathbb{R}^m$ embedded in $\mathbb{R}^d$, i.e., Property 1 and Property 2.

For property 1 (Nearly Unit Vectors), we need the following lemma:

**Lemma 1** *Given tree $\tau$, vector $\widehat{f}(\tau)$ has norm equal to 1.*

This lemma can be easily proven using property 2.4 and knowing that vectors in $\mathcal{N}$ are versors.

For property 2 (Nearly Orthogonal Vectors), we first need to observe that, due to properties 2.1 and 2.2, a tree $\tau$ generates a unique sequence of application of function $\diamond$ in $\widehat{f}(\tau)$ representing its structure. We can now address the following lemma:

**Lemma 2** *Given two different trees $\tau_a$ and $\tau_b$, the corresponding DTFs are nearly orthogonal: $|\widehat{f}(\tau_a) \cdot \widehat{f}(\tau_b)| < \epsilon$.*

**Proof** The proof is done by induction on the structure of $\tau_a$ and $\tau_b$.

**Basic step**

---

[1] We assume the degree of commutativity $k$ as the lowest number such that $\diamond$ is non-commutative, i.e., $\widetilde{\vec{a}} \diamond \widetilde{\vec{b}} \neq \widetilde{\vec{b}} \diamond \widetilde{\vec{a}}$, and for any $j < k$, $\widetilde{\vec{a}} \diamond c_1 \diamond \ldots \diamond c_j \diamond \widetilde{\vec{b}} \neq \widetilde{\vec{b}} \diamond c_1 \diamond \ldots \diamond c_j \diamond \widetilde{\vec{a}}$

Let $\tau_a$ be the single node $a$. Two cases are possible: $\tau_b$ is the single node $b \neq a$. Then, by the properties of vectors in $\mathcal{N}$, $|\widehat{f}(\tau_a) \cdot \widehat{f}(\tau_b)| = |\widetilde{\vec{a}} \cdot \widetilde{\vec{b}}| < \epsilon$; Otherwise, by Property 2.5, $|\widehat{f}(\tau_a) \cdot \widehat{f}(\tau_b)| = |\widetilde{\vec{a}} \cdot \widehat{f}(\tau_b)| < \epsilon$.

**Induction step**

***Case 1*** Let $\tau_a$ be a tree with root production $a \to a_1 \ldots a_k$ and $\tau_b$ be a tree with root production $b \to b_1 \ldots b_h$. The expected property becomes $|\widehat{f}(\tau_a) \cdot \widehat{f}(\tau_b)| = |(\widetilde{\vec{a}} \diamond \widehat{f}(\tau_{a_1} \ldots \tau_{a_k})) \cdot (\widetilde{\vec{b}} \diamond \widehat{f}(\tau_{b_1} \ldots \tau_{b_h}))| < \epsilon$. We have two cases: If $a \neq b$, $|\widetilde{\vec{a}} \cdot \widetilde{\vec{b}}| < \epsilon$. Then, $|\widehat{f}(\tau_a) \cdot \widehat{f}(\tau_b)| < \epsilon$ by Property 2.6. Else if $a = b$, then $\tau_{a_1} \ldots \tau_{a_k} \neq \tau_{b_1} \ldots \tau_{b_h}$ as $\tau_a \neq \tau_b$. Then, as $|\widehat{f}(\tau_{a_1} \ldots \tau_{a_k}) \cdot \widehat{f}(\tau_{b_1} \ldots \tau_{b_h})| < \epsilon$ is true by inductive hypothesis, $|\widehat{f}(\tau_a) \cdot \widehat{f}(\tau_b)| < \epsilon$ by Property 2.6.

***Case 2*** Let $\tau_a$ be a tree with root production $a \to a_1 \ldots a_k$ and $\tau_b = \tau_{b_1} \ldots \tau_{b_h}$ be a sequence of trees. The expected property becomes $|\widehat{f}(\tau_a) \cdot \widehat{f}(\tau_b)| = |(\widetilde{\vec{a}} \diamond \widehat{f}(\tau_{a_1} \ldots \tau_{a_k})) \cdot (\widehat{f}(\tau_{b_1}) \diamond \widehat{f}(\tau_{b_2} \ldots \tau_{b_h}))| < \epsilon$. Since $|\widetilde{\vec{a}} \cdot \widehat{f}(\tau_{b_1})| < \epsilon$ is true by inductive hypothesis, $|\widehat{f}(\tau_a) \cdot \widehat{f}(\tau_b)| < \epsilon$ by Property 2.6.

***Case 3*** Let $\tau_a = \tau_{a_1} \ldots \tau_{a_k}$ and $\tau_b = \tau_{b_1} \ldots \tau_{b_h}$ be two sequences of trees. The expected property becomes $|\widehat{f}(\tau_a) \cdot \widehat{f}(\tau_b)| = |(\widehat{f}(\tau_{a_1}) \diamond \widehat{f}(\tau_{a_2} \ldots \tau_{a_k})) \cdot (\widehat{f}(\tau_{b_1}) \diamond \widehat{f}(\tau_{b_2} \ldots \tau_{b_h}))| < \epsilon$. We have two cases: If $\tau_{a_1} \neq \tau_{b_1}$, $|\widehat{f}(\tau_a) \cdot \widehat{f}(\tau_b)| < \epsilon$ by inductive hypothesis. Then, $|\widehat{f}(\tau_a) \cdot \widehat{f}(\tau_b)| < \epsilon$ by Property 2.6. Else, if $\tau_{a_1} = \tau_{b_1}$, then $\tau_{a_2} \ldots \tau_{a_k} \neq \tau_{b_2} \ldots \tau_{b_h}$ as $\tau_a \neq \tau_b$. Then, as $|\widehat{f}(\tau_{a_2} \ldots \tau_{a_k}) \cdot \widehat{f}(\tau_{b_2} \ldots \tau_{b_h})| < \epsilon$ is true by inductive hypothesis, $|\widehat{f}(\tau_a) \cdot \widehat{f}(\tau_b)| < \epsilon$ by Property 2.6. ∎

### 3.2. Recursive Algorithm for Distributed Trees

This section discusses how to efficiently compute DTs. We focus on the space of tree fragments implicitly defined in (Collins & Duffy, 2001). This feature space refers to subtrees as any subgraph which includes more than one node, with the restriction that entire (not partial) rule productions must be included. We want to show that the related distributed trees can be recursively computed using a dynamic programming algorithm without enumerating the subtrees. We first define the recursive function and then we show that it exactly computes DTs.

### 3.2.1. RECURSIVE FUNCTION

The structural recursive formulation for the computation of distributed trees $\widetilde{\vec{T}}$ is the following:

$$\widetilde{\vec{T}} = \sum_{n \in N(T)} s(n) \qquad (6)$$

where $N(T)$ is the node set of tree $T$ and $s(n)$ represents the sum of distributed vectors for the subtrees of $T$ rooted in node $n$. Function $s(n)$ is recursively defined as follows:

- $s(n) = \vec{0}$ if $n$ is a terminal node.
- $s(n) = \widetilde{\vec{n}} \diamond (\widetilde{\vec{c_1}} + \sqrt{\lambda} s(c_1)) \diamond \ldots \diamond (\widetilde{\vec{c_m}} + \sqrt{\lambda} s(c_m))$ if $n$ is a node with children $c_1 \ldots c_m$.

As for the classic TK, the decay factor $\lambda$ decreases the weight of large tree fragments in the final kernel value. With dynamic programming, the time complexity of this function is linear $O(|N(T)|)$ and the space complexity is $d$ (where $d$ is the size of the vectors in $\mathbb{R}^d$).

### 3.2.2. THE RECURSIVE FUNCTION COMPUTES DISTRIBUTED TREES

The overall theorem we need is the following.

**Theorem 3** *Given the ideal vector composition function $\diamond$, the equivalence between equation (5) and equation (6) holds, i.e.:*

$$\widetilde{\vec{T}} = \sum_{n \in N(T)} s(n) = \sum_{\tau_i \in \mathcal{F}(T)} \omega_i \widehat{f}(\tau_i)$$

According to (Collins & Duffy, 2001), the contribution of tree fragment $\tau$ to the TK is $\lambda^{|\tau|-1}$, where $|\tau|$ is the number of nodes in $\tau$. Thus, we consider $\omega_i = \sqrt{\lambda^{|\tau_i|-1}}$. We demonstrate Theorem 3 by showing that $s(n)$ computes the weighted sum of vectors for the subtrees rooted in $n$ (see Theorem 5).

**Definition 3** *Let $n$ be a node of tree $T$. We define $R(n) = \{\tau | \tau \text{ is a subtree of } T \text{ rooted in } n\}$*

We need to introduce a simple lemma, whose proof is trivial.

**Lemma 4** *Let $\tau$ be a tree with root node $n$. Let $c_1, \ldots, c_m$ be the children of $n$. Then $R(n)$ is the set of all trees $\tau' = (n, \tau_1, ..., \tau_m)$ such that $\tau_i \in R(c_i) \cup \{c_i\}$.*

Now we can show that function $s(n)$ computes exactly the weighted sum of the distributed tree fragments for all the subtrees rooted in $n$.

**Theorem 5** *Let $n$ be a node of tree $T$. Then $s(n) = \sum_{\tau \in R(n)} \sqrt{\lambda^{|\tau|-1}} \widehat{f}(\tau)$.*

**Proof** The theorem is proved by structural induction.

**Basis.** Let $n$ be a terminal node. Then we have $R(n) = \emptyset$. Thus, by its definition, $s(n) = \vec{0} = \sum_{\tau \in R(n)} \sqrt{\lambda^{|\tau|-1}} \widehat{f}(\tau)$.

**Step.** Let $n$ be a node with children $c_1, \ldots, c_m$. The inductive hypothesis is then $s(c_i) = \sum_{\tau \in R(c_i)} \sqrt{\lambda^{|\tau|-1}} \widehat{f}(\tau)$.

Applying the inductive hypothesis, the definition of $s(n)$ and the property 2.3, we have

$$s(n) = \widetilde{\vec{n}} \diamond \left(\widetilde{\vec{c_1}} + \sqrt{\lambda} s(c_1)\right) \diamond \ldots \diamond \left(\widetilde{\vec{c_m}} + \sqrt{\lambda} s(c_m)\right)$$

$$= \widetilde{\vec{n}} \diamond \left(\widetilde{\vec{c_1}} + \sum_{\tau_1 \in R(c_1)} \sqrt{\lambda^{|\tau_1|}} \widehat{f}(\tau_1)\right) \diamond \ldots \diamond$$

$$\left(\widetilde{\vec{c_m}} + \sum_{\tau_m \in R(c_m)} \sqrt{\lambda^{|\tau_m|}} \widehat{f}(\tau_m)\right)$$

$$= \widetilde{\vec{n}} \diamond \sum_{\tau_1 \in \mathcal{T}_1} \sqrt{\lambda^{|\tau_1|}} \widehat{f}(\tau_1) \diamond \ldots \diamond \sum_{\tau_m \in \mathcal{T}_m} \sqrt{\lambda^{|\tau_m|}} \widehat{f}(\tau_m)$$

$$= \sum_{(n,\tau_1,\ldots,\tau_m) \in \{n\} \times \mathcal{T}_1 \times \ldots \times \mathcal{T}_m} \sqrt{\lambda^{|\tau_1|+\ldots+|\tau_m|}} \widetilde{\vec{n}} \diamond \widehat{f}(\tau_1) \diamond$$

$$\ldots \diamond \widehat{f}(\tau_m)$$

where $\mathcal{T}_i$ is the set $R(c_i) \cup \{c_i\}$. Thus, by means of Lemma 4 and the definition of $\widehat{f}$, we can conclude that $s(n) = \sum_{\tau \in R(n)} \sqrt{\lambda^{|\tau|-1}} \widehat{f}(\tau)$. ∎

## 4. Comparative Analysis of Computational Complexity

DTKs have an attractive constant computational complexity. We here compare their complexity with respect to the traditional tree kernels (TK) (Collins & Duffy, 2001), the fast tree kernels (FTK) (Moschitti, 2006), the fast tree kernels plus feature selection (FTK+FS) (Pighin & Moschitti, 2010), and the approximate tree kernels (ATK) (Rieck et al., 2010). We discussed basic features of these kernels in the introduction.

Table 4 reports time and space complexity of the kernels in *learning* and in *classification*. DTK is clearly competitive with respect to other methods, since both complexities are constant, according to the size $d$ of the reduced feature space. In these two phases, kernels are applied many times by the learning algorithms. Then, a constant complexity is extremely important. Clearly, there is a trade-off between the chosen $d$ and the average size of trees $n$. A comparison among execution times is done applying these algorithms to actual trees (see Section 5.3).

## 5. Empirical Analysis and Experimental Evaluation

In this section we propose two approximations of the ideal composition function $\diamond$, we investigate on their appropri-

**Distributed Tree Kernels**

| | Learning | | Classification | |
|---|---|---|---|---|
| | Time | Space | Time | Space |
| TK | $O(n^2)$ | $O(n^2)$ | $O(n^2)$ | $O(n^2)$ |
| FTK | $A(n)$ | $O(n^2)$ | $A(n)$ | $O(n^2)$ |
| FTK+FS | $A(n)$ | $O(n^2)$ | k | k |
| ATK | $O(\frac{n^2}{q_\omega})$ | $O(n^2)$ | $O(\frac{n^2}{q_\omega})$ | $O(n^2)$ |
| DTK | d | d | d | d |

Table 1. Computational time and space complexities for several tree kernel techniques: $n$ is the tree dimension, $q_\omega$ is a speed-up factor, $k$ is the size of the selected feature set, $d$ is the dimension of space $R^d$, $O(\cdot)$ is the worst-case complexity, and $A(\cdot)$ is the average case complexity.

ateness with respect to the ideal properties, we evaluate whether these concrete basic composition functions yield to effective DTKs, and, finally, we evaluate the computation efficiency by comparing average computational execution times of TKs and DTKs. For the following experiments, we focus on a reduced space $\mathbb{R}^d$ with $d = 8192$.

### 5.1. Approximating Ideal Basic Composition Function

#### 5.1.1. CONCRETE COMPOSITION FUNCTIONS

We consider two possible approximations for the ideal composition function $\diamond$: the *shuffled $\gamma$-product* $\boxtimes$ and *shuffled circular convolution* $\boxdot$. These functions are defined as follows:

$$\widetilde{\vec{a}} \boxtimes \widetilde{\vec{b}} = \gamma \cdot p_1(\widetilde{\vec{a}}) \otimes p_2(\widetilde{\vec{b}})$$
$$\widetilde{\vec{a}} \boxdot \widetilde{\vec{b}} = p_1(\widetilde{\vec{a}}) \odot p_2(\widetilde{\vec{b}})$$

where: $\otimes$ is the element-wise product between vectors and $\odot$ is the circular convolution (as for distributed representations in (Plate, 1995)) between vectors; $p_1$ and $p_2$ are two different permutations of the vector elements; and $\gamma$ is a normalization scalar parameter, computed as the average norm of the element-wise product of two vectors.

#### 5.1.2. EMPIRICAL EVALUATIONS OF PROPERTIES

Properties 2.1, 2.2, and 2.3 hold by construction. The two permutation functions, $p_1$ and $p_2$, guarantee Prop. 2.1, for a high degree $k$, and Prop. 2.2. Property 2.3 is inherited from element-wise product $\otimes$ and circular convolution $\odot$.

Properties 2.4, 2.5 and 2.6 can only be approximated. Thus, we performed tests to evaluate the appropriateness of the two considered functions.

Property 2.4 approximately holds for $\boxdot$ since approximate norm preservation already holds for circular convolution, whereas $\boxtimes$ uses factor $\gamma$ to preserve norm. We empirically evaluated this property. Figure 2(a) shows the average norm for the composition of an increasing number of basic vectors (i.e. vectors with unitary norm) with the two basic composition functions. Function $\boxdot$ behaves much better than $\boxtimes$.

Properties 2.5 and 2.6 were tested by measuring similarities between some combinations of vectors. The first experiment compared a single vector $\widetilde{\vec{a}}$ to a combination $\widetilde{\vec{t}}$ of several other vectors, as in property 2.5. Both functions resulted in average similarities below 1%, independently of the number of vectors in $\widetilde{\vec{t}}$, satisfying property 2.5. To test property 2.6 we compared two compositions of vectors $\widetilde{\vec{a}} \diamond \widetilde{\vec{t}}$ and $\widetilde{\vec{b}} \diamond \widetilde{\vec{t}}$, where all the vectors are in common except for the first one. The average similarity fluctuates around 0, with $\boxdot$ performing better than $\boxtimes$; this is mostly notable observing that the variance grows with the number of vectors in $\widetilde{\vec{t}}$ as shown in Fig. 2(b). A similar test was performed, with all the vectors in common except for the last one, yielding to similar results.

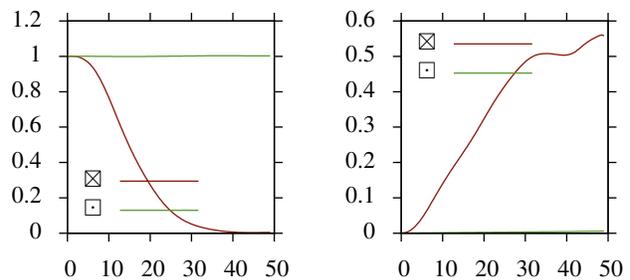

(a) Average norm of the vector obtained as combination of different numbers of basic random vectors

(b) Variance of the dot product between two combinations of basic random vectors with one common vector

Figure 2. Statistical properties for vectors on 100 samples ($d = 8192$).

In light of these results, $\boxdot$ seems to be a better choice than $\boxtimes$, although it should be noted that, for vectors of dimension $d$, $\boxtimes$ is computed in $O(d)$ time, while $\boxdot$ takes $O(d \log d)$ time.

### 5.2. Evaluating Distributed Tree Kernels: Direct and Task-based Comparison

In this section, we evaluate whether DTKs with the two concrete composition functions, $DTK_\boxtimes$ and $DTK_\boxdot$, approximate the original TK (as in Equation 4). We perform two sets of experiments: (1) a *direct comparison* where we directly investigate the correlation between DTK and TK values; and, (2) a *task based comparison* where we compare the performance of DTK against that of TK on two natural language processing tasks, i.e., question classification (QC) and textual entailment recognition (RTE).

#### 5.2.1. EXPERIMENTAL SET-UP

For the experiments, we used standard datasets for the two NLP tasks of QC and RTE.

| | QC | | RTE | |
|---|---|---|---|---|
| $\lambda$ | $DTK_\boxtimes$ | $DTK_\square$ | $DTK_\boxtimes$ | $DTK_\square$ |
| 0.2 | 0.993 | 0.994 | 0.997 | 0.998 |
| 0.4 | 0.980 | 0.989 | 0.990 | 0.961 |
| 0.6 | 0.908 | 0.880 | 0.890 | 0.350 |
| 0.8 | 0.644 | 0.377 | 0.469 | 0.039 |
| 1.0 | 0.316 | 0.107 | 0.169 | 0.000 |

*Table 2.* Spearman's correlation between DTK values and TK values. Test trees were taken from the QC corpus in table (a) and the RTE corpus in table (b).

For QC, we used a standard question classification training and test set[2], where the test set are the 500 TREC 2001 test questions. To measure the task performance, we used a question multi-classifier by combining $n$ binary SVMs according to the ONE-vs-ALL scheme, where the final output class is the one associated with the most probable prediction.

For RTE we considered the corpora ranging from the first challenge to the fifth (Dagan et al., 2006), except for the fourth, which has no training set. These sets are referred to as RTE1-5. The dev/test distribution for RTE1-3, and RTE5 is respectively 567/800, 800/800, 800/800, and 600/600 T-H pairs. We used these sets for the traditional task of pair-based entailment recognition, where a pair of text-hypothesis $p = (t, h)$ is assigned a positive or negative entailment class. For our comparative analysis, we use the syntax-based approach described in (Moschitti & Zanzotto, 2007) with two kernel function schemes: (1) $PK_S(p_1, p_2) = K_S(t_1, t_2) + K_S(h_1, h_2)$; and, (2) $PK_{S+Lex}(p_1, p_2) = Lex(t_1, h_1)Lex(t_2, h_2) + K_S(t_1, t_2) + K_S(h_1, h_2)$. $Lex$ is a standard similarity feature between the text and the hypothesis and $K_S$ is realized with $TK$, $DTK_\boxtimes$, and $DTK_\square$. In the plots, the different $PK_S$ kernels are referred to as $TK$, $DTK_\boxtimes$, and $DTK_\square$ whereas the different $PK_{S+Lex}$ kernels are referred to as $TK + Lex$, $DTK_\boxtimes + Lex$, and $DTK_\square + Lex$.

### 5.2.2. CORRELATION BETWEEN TK AND DTK

As a first measure of the ability of DTK to emulate the classic TK, we considered the Spearman's correlation of their values computed on the parse trees for the sentences contained in QC and RTE corpora. Table 2 reports results and shows that DTK does not approximate adequately TK for $\lambda = 1$. This highlights the difficulty of DTKs to correctly handle pairs of large *active forests*, i.e., trees with many subtrees with weights around 1. The correlation improves dramatically when parameter $\lambda$ is reduced. We can conclude that DTKs efficiently approximate TK for the $\lambda \leq 0.6$. These values are relevant for the applications as we will also see in the next section.

---

[2]The QC set is available at http://l2r.cs.uiuc.edu/~cogcomp/Data/QA/QC/

### 5.2.3. TASK-BASED COMPARISON

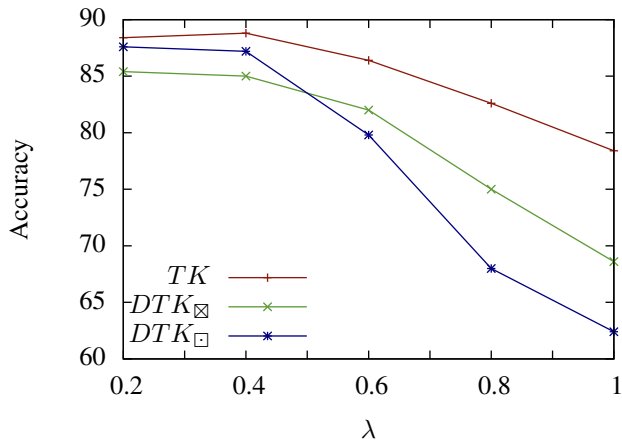

*Figure 3.* Performance on Question Classification task ($DTK_\boxtimes$ and $DTK_\square$ rely on vectors of $d = 8192$).

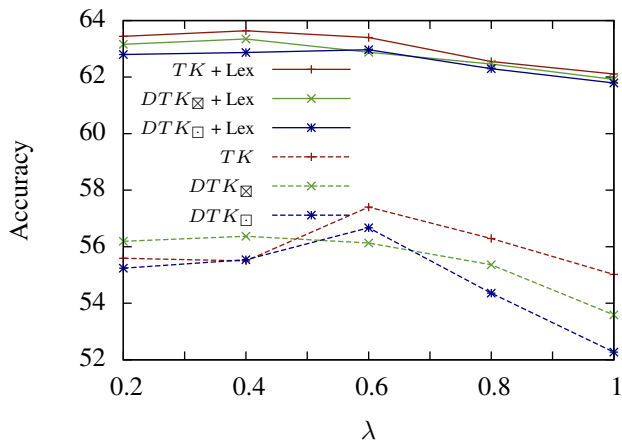

*Figure 4.* Performance on Recognizing Textual Entailment task ($DTK_\boxtimes$ and $DTK_\square$ rely on vectors of $d = 8192$). Each point is the average of accuracy on the 4 data sets.

We performed both QC and RTE experiments for different values of parameter $\lambda$. Results are shown in Fig. 3 and 4 for QC and RTE tasks respectively.

For QC, DTK leads to worse performances with respect to TK, but the gap is narrower for small values of $\lambda \leq 0.4$ (with $DTK_\square$ better than $DTK_\boxtimes$). These $\lambda$ values produce better performance for the task. For RTE, for $\lambda \leq 0.4$, $DTK_\boxtimes$ and $DTK_\square$ is similar to $TK$. Differences are not statistically significant except for for $\lambda = 0.4$ where $DTK_\boxtimes$ behaves better than $TK$ (with $p < 0.1$). Statistical significance is computed using the two-sample Student t-test. $DTK_\boxtimes + Lex$ and $DTK_\square + Lex$ are statisti-cally similar to $TK + Lex$ for any value of $\lambda$. DTKs are a good approximation of TKs for $\lambda \leq 0.4$, that are the values where $TK$s have the best performances in the tasks.

### 5.3. Average computation time

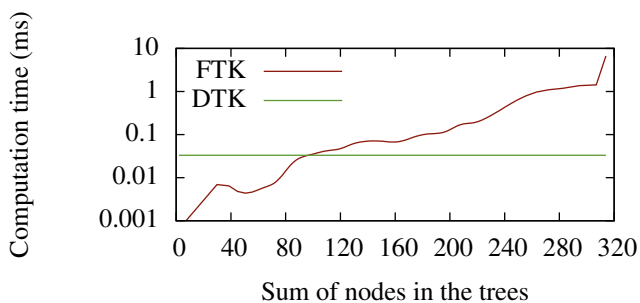

*Figure 5.* Computation time of FTK and DTK (with $d = 8192$) for tree pairs with an increasing total number of nodes, on a 1.6 GHz CPU.

We measured the average computation time of FTK (Moschitti, 2006) and DTK (with vector size 8192) on trees from the Question Classification corpus. Figure 5 shows the relation between the computation time and the size of the trees, computed as the total number of nodes in the two trees. As expected, DTK has constant computation time, since it is independent of the size of the trees. On the other hand, computation time for FTK, while being lower for smaller trees, grows very quickly with the tree size. The larger are the trees considered, the higher is the computational advantage offered by using DTK instead of FTK.

## 6. Conclusion

In this paper we proposed the distributed tree kernels (DTKs) as an approach to reduce computational complexity of tree kernels. Having an ideal function for vector composition, we have formally shown that high-dimensional spaces of tree fragments can be embedded in low-dimensional spaces where tree kernels can be directly computed with dot products. We have empirically shown that we can approximate the ideal function for vector composition. The resulting DTKs correlate with original tree kernels, obtain similar results in two natural language processing tasks, and, finally, are faster.